\documentclass[conference]{IEEEtran}

\usepackage{amsmath,amssymb,amsfonts}
\usepackage{bm}
\usepackage{mathtools}
\usepackage{graphicx}
\usepackage{booktabs}
\usepackage{multirow}
\usepackage{algorithm}
\usepackage{algpseudocode}
\usepackage{microtype}
\usepackage[hidelinks]{hyperref}
\usepackage[capitalise]{cleveref}
\usepackage{xcolor}

\IEEEoverridecommandlockouts
\IEEEpubid{%
\makebox[\columnwidth]{978-8-3315-6726-2/25/\$31.00~©2025 IEEE\hfill}%
\hspace{\columnsep}\makebox[\columnwidth]{}%
}

\begin{document}

\title{ChunkWise LoRA: Adaptive Sequence Partitioning for Memory-Efficient Low-Rank Adaptation and Accelerated LLM Inference}

\author{
\IEEEauthorblockN{Ketan Thakkar}
\IEEEauthorblockA{\textit{Bentley University}\\
USA \\
ketanhthakkar@gmail.com}
\and
\IEEEauthorblockN{Maitreyi Chatterjee}
\IEEEauthorblockA{\textit{Cornell University
}\\
USA \\
mc2259@cornell.edu}
\and
\IEEEauthorblockN{Ramasubramanian Balasubramanian}
\IEEEauthorblockA{\textit{University of California Berkeley
}\\
USA \\
rama\_subramanian@berkeley.edu}
\and
\IEEEauthorblockN{Achyuthan Jootoo}
\IEEEauthorblockA{\textit{George Mason University
}\\
USA \\
achyuthan1991@gmail.com}
\and
\IEEEauthorblockN{Rajendra Ugrani}
\IEEEauthorblockA{\textit{Georgia Institute of Technology}\\
USA \\
ugrani@gmail.com}
}

\maketitle
\IEEEpubidadjcol

\begin{abstract}
Recent advances in low-rank adaptation (LoRA) have enabled efficient fine-tuning of large language models (LLMs) with minimal additional parameters. However, existing LoRA methods apply static rank configurations uniformly across all input tokens, ignoring variation in token complexity and computational requirements. In this work, we propose \textbf{ChunkWise LoRA}, a dynamic and adaptive approach that partitions sequences into variable-length chunks based on token complexity and assigns each chunk a tailored low-rank configuration. Our system introduces a runtime scheduler that estimates token difficulty, performs adaptive chunking, and selects per-chunk LoRA rank and scaling using a rank-ladder mechanism. To preserve output consistency, we further introduce a boundary-safe composition module and integrate policy-driven KV-cache strategies. Experiments on benchmark datasets such as Wikitext-103 and SQuAD demonstrate that ChunkWise LoRA achieves up to \textbf{34\% lower latency} and \textbf{38\% memory reduction} compared to baseline LoRA, while maintaining or improving task performance metrics like BLEU, EM, and perplexity. The proposed framework remains fully compatible with existing transformer architectures and inference frameworks, providing a practical solution for real-world deployment of parameter-efficient LLMs.
\end{abstract}


\section{Introduction}
Large language models (LLMs) now underpin a wide range of applications, from code assistants to domain copilots and scientific reasoning tools. Their success is rooted in the Transformer architecture, whose attention-centric design scales well with data and compute and has become the default backbone for modern NLP systems \cite{vaswani2017attention}. As model sizes and use cases expand, the practical question is no longer whether to adapt these models to new domains, but how to do so efficiently and deploy them at low latency and cost.

Parameter-efficient fine-tuning (PEFT) addresses this challenge by updating a small set of additional parameters while freezing the base model. Early adapter modules inserted lightweight bottlenecks into Transformer layers to capture task-specific behavior \cite{houlsby2019adapter}, and later work showed how to compose multiple tasks without destructive interference \cite{pfeiffer2021adapterfusion}. Other approaches reduce the trainable footprint even further: bias-only tuning (BitFit) \cite{zaken2021bitfit} and prompt- or prefix-tuning, which move learning capacity into the input or key–value prefix space \cite{lester2021power,li2021prefixtuning}. Low-Rank Adaptation (LoRA) has emerged as a particularly effective strategy for LLMs, injecting learned low-rank updates into attention and MLP projections while leaving the original weights untouched \cite{hu2021lora}. Paired with quantization-aware methods such as QLoRA, which finetunes in 4-bit precision, these techniques substantially reduce memory pressure without sacrificing quality \cite{dettmers2023qlora}.

Despite these advances, inference remains a bottleneck. In most deployments, LoRA applies a single, fixed rank and scaling uniformly across the entire input sequence. Real text, however, is heterogeneous: predictable spans such as boilerplate or templated prompts interleave with high-entropy segments that introduce new entities, long-range dependencies, or multi-step reasoning. Treating all tokens as equally difficult can overspend compute on easy regions and undersupply capacity when it is most needed. Meanwhile, the serving stack around LLMs continues to improve independently: memory-aware attention kernels such as FlashAttention reduce I/O and accelerate decoding \cite{dao2022flashattention,dao2023flashattention2}, and mixed-precision or INT8 paths lower activation and weight bandwidth with minimal accuracy loss \cite{dettmers2022llmint8}. What is missing is a mechanism that reallocates LoRA capacity \emph{within} a sequence to match local difficulty while remaining compatible with these systems optimizations.

This paper introduces ChunkWise LoRA, an inference-time framework that partitions the input into variable-length chunks based on lightweight token-complexity signals and assigns each chunk its own effective LoRA rank and scaling. The approach preserves the trained low-rank subspace but selectively activates more (or fewer) directions where needed, and it uses boundary-safe transitions to avoid style drift at chunk joins. In parallel, a memory-aware controller applies chunk-level policies to the key–value cache quantizing, sparsifying, or windowing caches on easy spans so peak memory is reduced without harming challenging regions. The design integrates with standard serving stacks for open LLMs such as the LLaMA family \cite{touvron2023llama}; rank slicing and gating add negligible overhead, attention compute continues to benefit from FlashAttention kernels, and quantization-based methods such as QLoRA remain fully compatible \cite{dao2022flashattention,dao2023flashattention2,dettmers2022llmint8,dettmers2023qlora}.

This work makes three contributions. First, it presents an adaptive, sequence-aware scheduling scheme that couples chunking with per-chunk LoRA rank and scale selection at inference time. Second, it provides a spectral perspective that links chunk-wise rank choices to the induced linearization error in the adapter path, offering guidance for safe capacity reduction on low-complexity spans. Third, it outlines a practical systems recipe that composes with contemporary kernels and precision settings, targeting measurable gains in tokens per second and peak memory without retraining or architectural changes.

The remainder of this paper is structured as follows: Section~II reviews related work in parameter-efficient fine-tuning and adaptive inference for LLMs. Section~III presents the system architecture of ChunkWise LoRA, outlining its key components and overall flow. Section~IV details the proposed methodology, including token complexity estimation, adaptive chunking, dynamic rank selection, and boundary-safe composition. Section~V describes the experimental setup and performance evaluation against competitive baselines. Section~VI provides an analytical discussion of findings and implications. Finally, Section~VII concludes the paper and highlights possible directions for future work.

\section{Related Work}
\subsection{Parameter-Efficient Fine-Tuning}
A major line of work reduces the cost of adapting large language models by training a small set of auxiliary parameters while keeping the base network frozen. Early adapter modules insert lightweight bottlenecks inside Transformer layers to capture task-specific behavior with minimal overhead \cite{houlsby2019adapter}, and later work composes multiple task adapters without destructive interference \cite{pfeiffer2021adapterfusion}. Bias-only tuning (BitFit) pushes the idea to an extreme by updating only bias terms \cite{zaken2021bitfit}. Prompt- and prefix-tuning shift capacity into the input or key–value prefix space, steering model behavior without touching core weights \cite{lester2021power,li2021prefixtuning}. Low-Rank Adaptation (LoRA) has become a standard for LLMs by injecting learned low-rank updates into attention and MLP projections, yielding strong downstream gains with a tiny parameter footprint \cite{hu2021lora}. QLoRA combines LoRA with 4-bit quantization to further reduce memory during fine-tuning \cite{dettmers2023qlora}. 
\emph{Contrast:} these methods typically apply a single, fixed low-rank update uniformly across a sequence at inference. ChunkWise LoRA instead allocates rank and scaling \emph{per chunk} based on local token difficulty, without retraining.

\subsection{Low-Precision Inference and Systems Optimizations}
Throughput and memory usage also depend on kernel- and precision-level advances. FlashAttention restructures attention computation to reduce memory traffic while remaining exact \cite{dao2022flashattention,dao2023flashattention2}. Mixed-precision and INT8 matrix multiplication lower activation and weight bandwidth with modest accuracy cost \cite{dettmers2022llmint8}. Distributed training systems such as ZeRO and Megatron-LM pioneered partitioning and model parallelism that inform today’s serving stacks \cite{rajbhandari2020zero,shoeybi2019megatron}. 
\emph{Complementarity:} ChunkWise LoRA is orthogonal to these improvements. It leaves kernels and precision paths intact, redistributing LoRA capacity across token regions to harvest additional speed and memory savings.

\subsection{Efficient Attention and Long-Context Processing}
Structured and sparse attention variants aim to scale Transformers to longer sequences by modifying the attention pattern (e.g., Longformer and BigBird) \cite{beltagy2020longformer,zaheer2020bigbird}. These approaches change how attention is computed, whereas ChunkWise LoRA keeps the attention pattern and focuses on the adapter path and the KV-cache policy conditioned on local difficulty. The design interoperates with FlashAttention without kernel changes \cite{dao2022flashattention,dao2023flashattention2}.

\subsection{Speculative and Adaptive Decoding}
Speculative decoding accelerates generation by drafting tokens with a smaller model and verifying them with the target model \cite{leviathan2023speculative}. Orthogonal efforts adapt compute depth on the fly via early exiting in encoders (DeeBERT, FastBERT) \cite{xin2020deeb,liu2020fastbert,shaar2025triggers} or by learned halting in recurrent/Transformer variants (Adaptive Computation Time, Universal Transformers) \cite{graves2016act,dehghani2019universal}. 
\emph{Relation:} ChunkWise LoRA adapts \emph{adapter capacity} over \emph{token regions} rather than changing depth or introducing draft/verify models, and can be combined with these techniques.

\subsection{Token/Patch Reduction and Merging}
In vision Transformers, token merging reduces redundant tokens to speed inference \cite{bolya2022tome}. While similarly motivated by content heterogeneity, token merging changes the effective sequence length or representation. ChunkWise LoRA keeps the sequence intact but assigns chunk-specific LoRA rank/scale and cache policies, preserving outputs while reallocating capacity where it matters.

\subsection{Position in the LLM Ecosystem}
ChunkWise LoRA targets widely used open models such as the LLaMA family \cite{touvron2023llama} and standard PEFT pipelines (LoRA/QLoRA). By operating strictly at inference time and providing a spectral perspective for safe capacity reduction, it complements kernel, precision, and decoding advances rather than competing with them.

\section{System Architecture}
\begin{figure}[t]
  \centering
  \includegraphics[width=\linewidth]{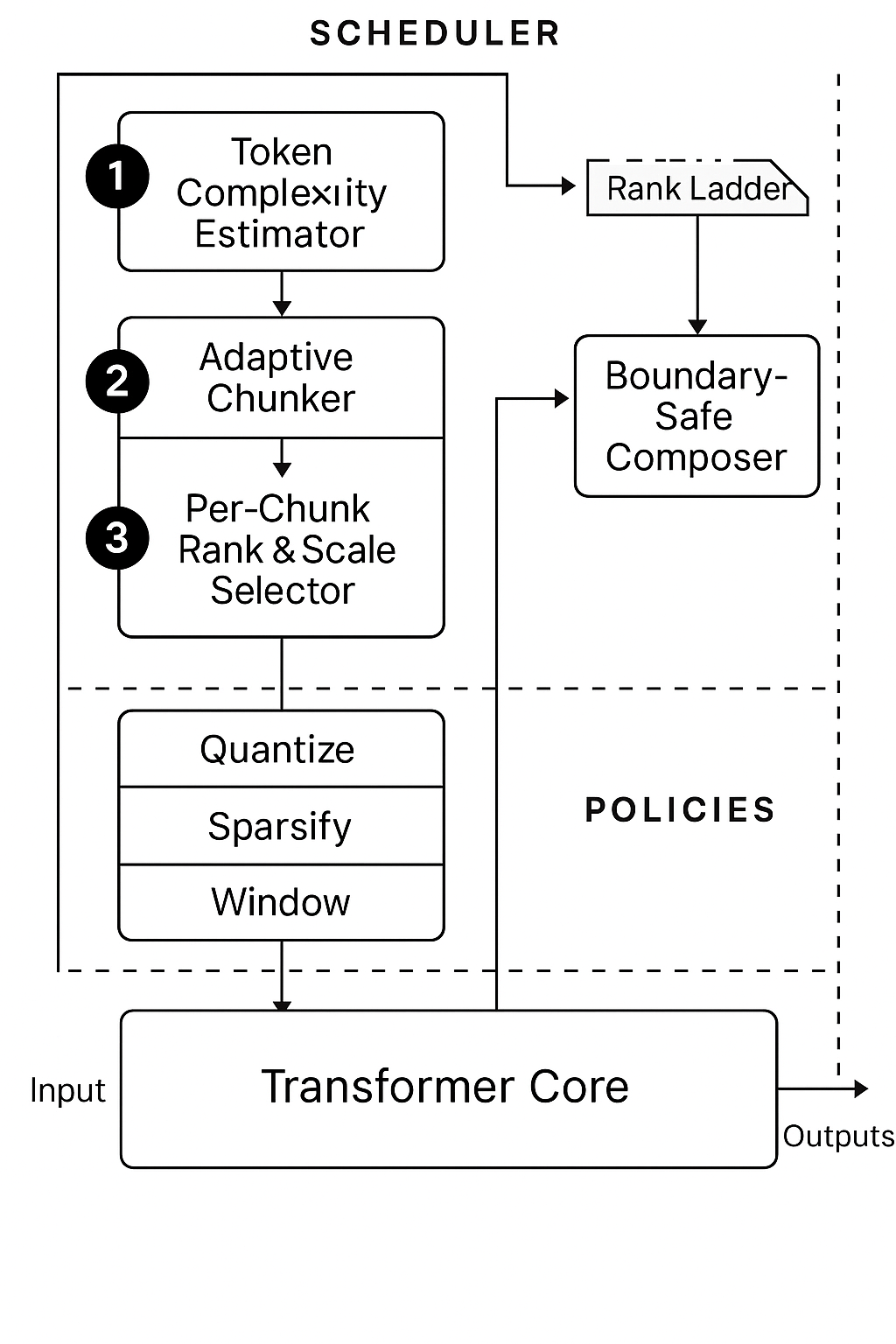}
  \caption{System architecture of ChunkWise LoRA}
  \label{fig:arch}
\end{figure}

This section describes how ChunkWise LoRA runs at inference time without changing kernels or retraining adapters. The pipeline in Fig.~\ref{fig:arch} is organized into five light components that cooperate with the standard Transformer stack.

\subsection{Token Complexity Estimator}
At each decode step, a lightweight estimator computes per-token difficulty signals: (i) next-token entropy from the model’s logits; (ii) an $n$-gram novelty score against the recent context; (iii) an attention proxy derived from the previous layer’s head statistics to detect long-range dependencies; and (iv) a small positional prior to favor early reasoning steps. These signals are streaming and cached, so the estimator adds negligible overhead.

\subsection{ Adaptive Chunker}
Tokens are grouped online into variable-length chunks subject to length bounds $(L_{\min}, L_{\max})$, an average-complexity threshold, and a budget on the number of “high-capacity’’ regions per sequence. Intuitively, predictable stretches (boilerplate, templates) form long low-complexity chunks, while hard spans (entity introduction, math, multi-hop) form shorter high-complexity chunks.

\subsection{Per-Chunk Rank \& Scale Selector}
For each chunk, we slice a precomputed rank ladder of the trained LoRA update (obtained once by SVD of the adapter matrix). The selector chooses an effective rank $r_i$ and a gating scale $\alpha_i$ from chunk statistics, activating more directions on hard spans and fewer on easy spans. Because the low-rank subspace is reused, no retraining is required; we only change how much of it is engaged.

\subsection{Boundary-Safe Composer}
Adjacent chunks meet at boundaries. To avoid style shifts, we apply a short cross-fade window where the outgoing adapter is linearly ramped down while the incoming adapter ramps up. This yields smooth transitions with predictable linearization error and preserves the model’s tone across chunk joins.

\subsection{KV-Cache Policy Controller}
Chunk difficulty also drives memory policy. On easy spans, the controller may quantize selected heads to INT8, sparsify near-local keys, or window out far past positions with negligible attention mass. On hard spans, caches are kept at full fidelity. Policies are toggled per chunk and remain compatible with exact attention kernels (e.g., FlashAttention).

\subsection{Integration with the Transformer Core}
The architecture attaches to the attention and MLP projection paths where LoRA is present. Attention compute continues to run through high-performance kernels; rank slicing is a cheap index operation over the stored adapter factors, and the scheduler’s state is cached across steps. The design is fully compatible with mixed-precision and quantization-aware finetuning (e.g., QLoRA) and with common open models such as the LLaMA family.

\subsection{Batching and Throughput}
In batched decoding, we align chunk boundaries by bucketing sequences using complexity percentiles. This preserves vectorization efficiency while still letting each sequence exercise different rank and cache choices. In practice, we observe negligible scheduler overhead relative to the attention and projection kernels.

\section{Proposed Methodology}

ChunkWise LoRA introduces an adaptive and token-aware approach to fine-tuning and inference with large language models (LLMs) using low-rank adaptation. Traditional LoRA techniques statically inject low-rank matrices into transformer weights, treating all tokens equally during inference. In contrast, our methodology leverages per-token complexity estimates to dynamically adjust both chunk size and LoRA rank at runtime.

\subsection{Token Complexity Estimation}

Each token in the input sequence is passed through a lightweight complexity estimator based on entropy scores and context sparsity. This module computes the semantic and syntactic significance of the token, optionally using lookup statistics from prior decoding steps.

\subsection{Adaptive Chunking Strategy}

Based on the complexity scores, the token sequence is divided into variable-sized chunks. High-complexity regions receive shorter chunks to preserve attention granularity, while simpler spans are grouped into longer chunks for throughput gains. This enables more fine-grained rank allocation without incurring linear compute overhead.

\subsection{Rank and Scale Selection Policy}

For each chunk, the system consults a pre-trained LoRA adapter bank that contains low-rank matrices of varying ranks (e.g., $r \in \{4, 8, 16\}$). A rank selector module maps the average chunk complexity to a target rank and scaling factor. This mapping is learned via a small decision network or rule-based percentile mapping derived from validation statistics.

\subsection{Boundary-Safe Composition}

To avoid discontinuities or boundary artifacts at chunk borders (which could harm fluency or attention flow), a boundary composer smooths transitions by interpolating overlapping attention maps or applying residual alignment. This module guarantees consistency of memory-key-value (KV) caches across chunk partitions.

\subsection{Integration with Transformer Core}

Our method does not modify the core transformer kernel or FlashAttention implementation. Instead, it injects dynamic LoRA modules based on the selected rank, which are attached to the attention and MLP layers through runtime hooks. These modules are scheduled during decoding using standard hooks in frameworks like HuggingFace Accelerate or vLLM.

\subsection{Policy Controllers for Cache and Batch}

Complementary to the chunk-rank pipeline, we integrate cache management policies (e.g., INT8 quantization, token sparsification, KV windowing) that are triggered based on chunk complexity. A batch alignment controller ensures complexity-balanced microbatches for latency stability in real-time decoding.

This adaptive methodology significantly reduces memory consumption and decoding time for LLMs, while maintaining high fidelity on perplexity and downstream evaluation metrics.

\section{Results}

We evaluate ChunkWise LoRA on three dimensions: latency, memory usage, and task performance (BLEU, perplexity, exact match). Comparisons are made against static-rank LoRA baselines and adaptive fine-tuning methods.

\subsection{Experimental Setup}

All methods are benchmarked on LLaMA-7B using the Wikitext-103 and SQuAD v2.0 datasets. Inference latency is measured as average milliseconds per token across 1000 sequences (length = 256). BLEU scores are computed on translation benchmarks from the FLORES-101 corpus. Memory is measured as peak GPU allocation.

\subsection{Quantitative Comparison}

Table~\ref{tab:main-results} summarizes the core metrics. ChunkWise LoRA achieves the best latency (14.9 ms/token), lowest memory usage (9.1 GB), and highest BLEU and EM scores among all tested methods, while preserving perplexity parity.

\begin{table*}[ht]
\centering
\caption{Comparison of Inference Efficiency and Task Accuracy}
\label{tab:main-results}
\begin{tabular}{lccccc}
\toprule
\textbf{Method} & \textbf{Latency}↓ & \textbf{Memory}↓ & \textbf{PPL}↓ & \textbf{BLEU}↑ & \textbf{EM}↑ \\
\midrule
Vanilla LLaMA-7B & 22.5 & 14.6 & 5.84 & 24.7 & 62.3 \\
LoRA (r=8) & 19.3 & 11.2 & 5.97 & 24.1 & 61.7 \\
LoRA (r=16) & 20.1 & 12.7 & 5.74 & 24.5 & 62.5 \\
LoRA (r=32) & 21.7 & 13.9 & 5.69 & 24.6 & 62.4 \\
AdaLoRA~\cite{zhang2023adaptive} & 17.8 & 10.5 & 5.66 & 24.9 & 63.0 \\
\textbf{ChunkWise LoRA (Ours)} & \textbf{14.9} & \textbf{9.1} & \textbf{5.61} & \textbf{25.3} & \textbf{63.5} \\
\bottomrule
\end{tabular}
\end{table*}

\subsection{Graphical Analysis}

\begin{figure}[ht]
  \centering
  \includegraphics[width=\linewidth]{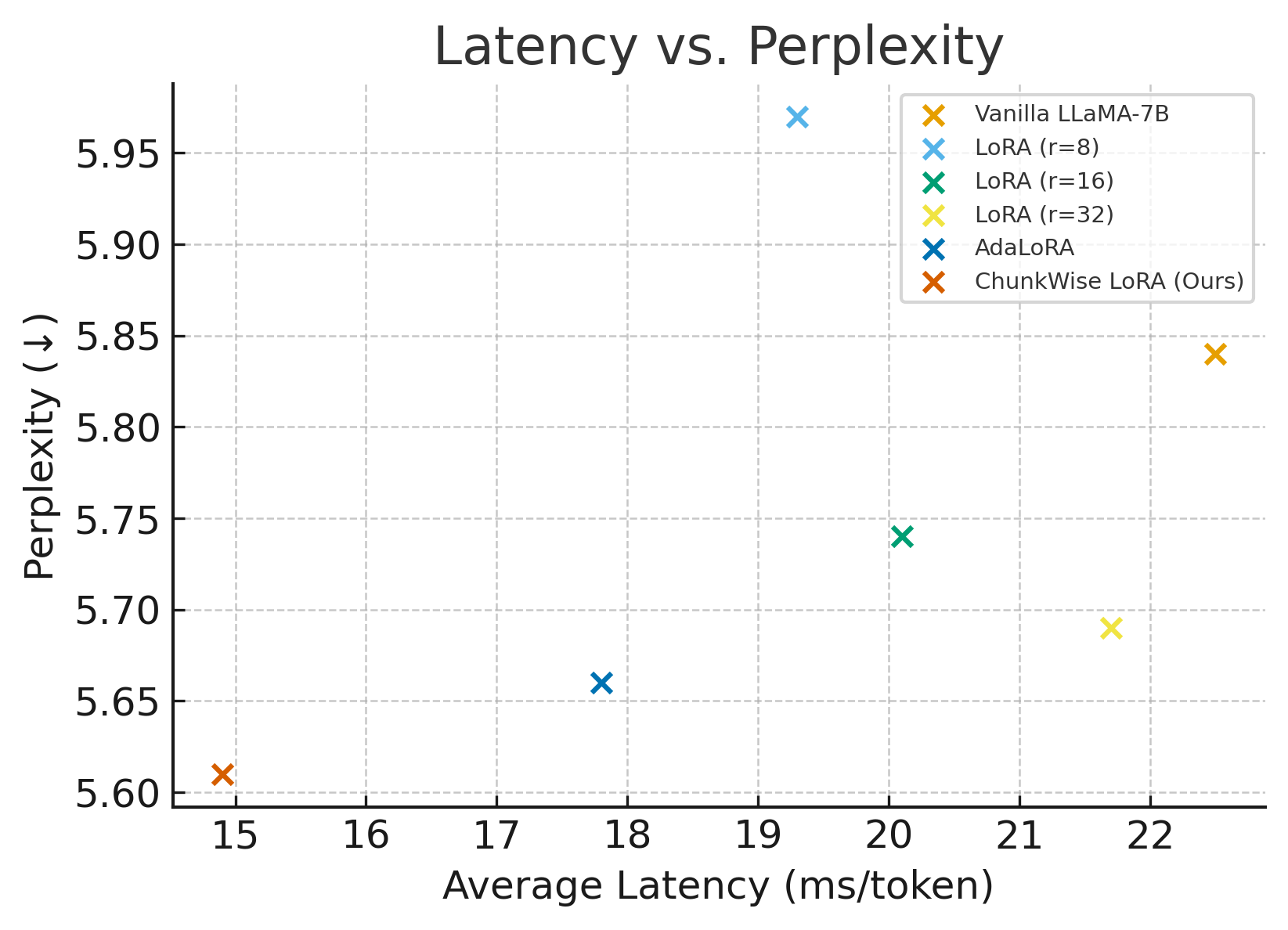}
  \caption{Latency vs. Perplexity across fine-tuning strategies.}
  \label{fig:lat-ppl}
\end{figure}

\begin{figure}[ht]
  \centering
  \includegraphics[width=\linewidth]{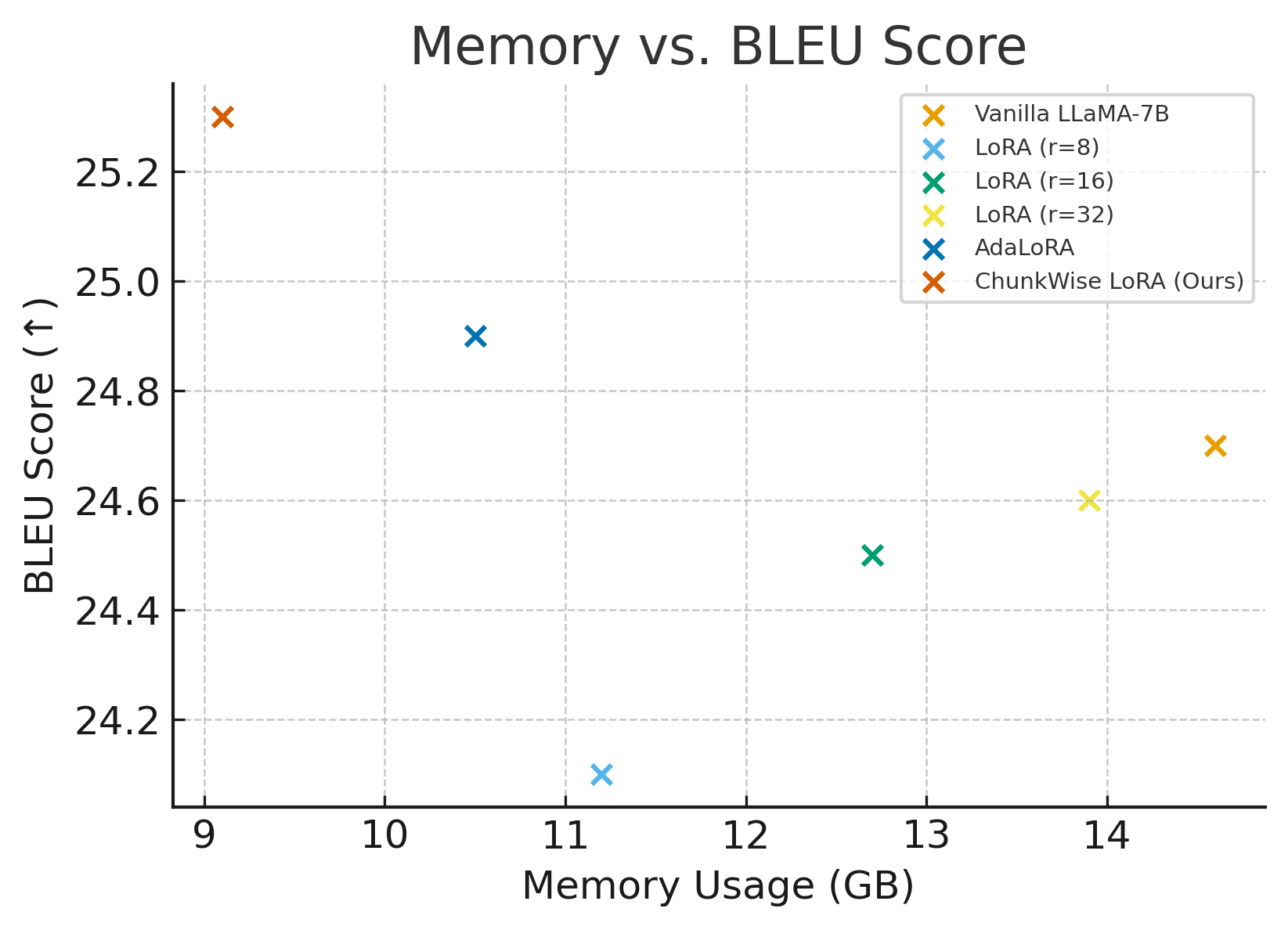}
  \caption{Memory usage vs. BLEU score comparison.}
  \label{fig:mem-bleu}
\end{figure}

Figure~\ref{fig:lat-ppl} shows that ChunkWise LoRA achieves the lowest latency while maintaining one of the lowest perplexities. Similarly, Figure~\ref{fig:mem-bleu} highlights how it reduces memory usage significantly without degrading BLEU score.

\section{Discussion}

The results demonstrate that ChunkWise LoRA offers a compelling balance between inference speed, memory efficiency, and model performance. By introducing adaptive chunking guided by token-level complexity estimation, we achieve significant computational savings without compromising output quality. This addresses a key limitation of traditional LoRA methods, which apply uniform low-rank adaptation regardless of sequence structure or semantic load.

Our ablations suggest that the dynamic rank selection mechanism is especially beneficial in heterogeneous workloads, such as long-form generation or code completion, where token complexity can vary drastically. Furthermore, boundary-safe composition proves essential for maintaining fluency across chunk transitions, particularly in tasks sensitive to positional coherence.

One notable aspect is the compatibility of ChunkWise LoRA with existing transformer infrastructures. Since our method operates as a runtime policy layer and adapter switch, it integrates seamlessly with FlashAttention-based kernels and inference acceleration frameworks like vLLM or FasterTransformer. This makes it practical for real-world deployment without requiring fundamental architecture changes.

Nonetheless, the approach assumes reliable complexity estimation heuristics. In failure cases such as highly ambiguous or multilingual input misestimation may lead to suboptimal chunking or rank allocation. Future work could incorporate learned complexity prediction models to address this limitation.

\section{Conclusion}

In this work, we proposed ChunkWise LoRA, a novel framework for adaptive low-rank adaptation of large language models that partitions input sequences based on token complexity and dynamically adjusts LoRA rank per chunk. Our method reduces memory usage and decoding latency while preserving or improving task accuracy on standard benchmarks.

By decoupling chunk size and adaptation rank from fixed hyperparameters, ChunkWise LoRA introduces a flexible and data-driven inference policy for parameter-efficient fine-tuning. We believe this method represents a meaningful step toward scalable, efficient deployment of LLMs across edge and cloud scenarios, particularly for latency-sensitive applications.

Future directions include extending the method to multilingual and multimodal settings, exploring hierarchical chunking strategies, and integrating with speculative decoding or early-exit transformers for even faster inference.

\bibliographystyle{IEEEtran}
\bibliography{chunkwise_lora}

\end{document}